\documentclass{article}

\usepackage{arxiv}

\usepackage{setspace}
\usepackage[utf8]{inputenc} 
\usepackage[T1]{fontenc}    
\usepackage[colorlinks=true, linkcolor=blue]{hyperref}
\usepackage{url}            
\usepackage{booktabs}       
\usepackage{amsfonts}       
\usepackage{microtype}      
\usepackage{graphicx}
\usepackage{doi}
\usepackage{minitoc}

\usepackage{subfigure}
\usepackage{color}
\usepackage{fancyhdr}
\usepackage[numbers]{natbib}

\usepackage{times}
\usepackage{amssymb, amsmath, amsbsy}
\usepackage{makecell}  
\usepackage{graphicx}
\usepackage{float}
\usepackage{multirow}
\usepackage{xcolor}
\usepackage{soul}
\usepackage{cleveref}
\usepackage{footmisc}

\newcommand{\bs}[1]{\boldsymbol{#1}}
\newcommand{\mc}[1]{\mathcal{#1}}

\DeclareMathOperator*{\argmin}{arg\,min}

\title{Deep learning for solution and inversion of structural mechanics and vibrations \thanks{A preprint to be published in: \emph{Modeling and Computations in Vibration Problems}, Edited by S.~Chakraverty, F.~Tornabene, J. N. ~Reddy, Institute of Physics Publishing. 2021. }}

\author{ \href{https://orcid.org/0000-0003-2659-0507}{\includegraphics[scale=0.06]{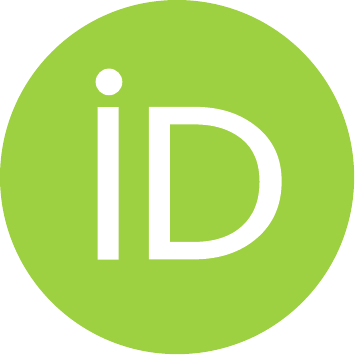}\hspace{1mm}Ehsan~Haghighat} \\
	Department of Civil Engineering\\
	University of British Columbia\\
	Vancouver, BC \\
	\texttt{ehsan.haghighat@ubc.ca} \\
	\And
	\href{https://orcid.org/0000-0002-9558-2087}{\includegraphics[scale=0.06]{orcid.pdf}\hspace{1mm}Ali C.~Bekar} \\
	Department of Aerospace and Mechanical Engineering\\
	University of Arizona\\
	Tucson, AZ \\
	\texttt{acbekar@email.arizona.edu} \\
	\And
	\href{https://orcid.org/0000-0003-2895-6111}{\includegraphics[scale=0.06]{orcid.pdf}\hspace{1mm}Erdogan~Madenci} \\
	Department of Aerospace and Mechanical Engineering\\
	University of Arizona\\
	Tucson, AZ \\
	\texttt{madenci@arizona.edu} \\
	\And
	\href{https://orcid.org/0000-0002-7370-2332}{\includegraphics[scale=0.06]{orcid.pdf}\hspace{1mm}Ruben~Juanes} \\
	Department of Civil and Environment Engineering\\
	Massachusetts Institute of Technology\\
	Boston, MA \\
	\texttt{juanes@mit.edu} \\
}

\renewcommand{\headeright}{ }
\renewcommand{\undertitle}{ }
\renewcommand{\shorttitle}{Deep Learning in Civil Engineering}

\hypersetup{
pdftitle={Deep learning for solution and inversion of structural mechanics and vibrations},
pdfsubject={q-bio.NC, q-bio.QM},
pdfauthor={Ehsan~Haghighat, Ali C.~Bekar, Erdogan~Madenci, Ruben~Juanes},
pdfkeywords={Deep learning, Physics-informed, Structural mechanics, Vibrations},
}

\date{October 2020}

\begin{document}
\maketitle

\begin{abstract}
	Deep learning has been the most popular machine learning method in the last few years. In this chapter, we present the application of deep learning and physics-informed neural networks concerning structural mechanics and vibration problems. Demonstration problems involve de-noising data, solution to time-dependent ordinary and partial differential equations, and characterizing the system's response for a given data.  
\end{abstract}



\section{Introduction}

The Finite Element Method (FEM) is commonly employed to construct and analyze a representative model of the structure and as a part of the inversion and identification techniques.  Inversion algorithms minimize the difference between the experimental data and computational model through direct displacement, mode shapes and, mode frequencies \cite{hall1970,kabe1985,ahmadian1997,friswell1998}. Classical studies primarily concern the identification of mechanical properties of the system such as stiffness and damping by matching the numerical predictions with experimental measurements. Recently, more fundamental statistical techniques such as Kalman filter and recurrent neural networks are used to identify the system and approximate its response \cite{yan2017, zhang2019}

Deep learning (DL), a subclass of machine learning (ML) and artificial intelligence (AI), has been in the forefront of recent advances in AI for addressing challenging problems in computer vision and autonomous driving, speech recognition and natural language processing, and e-commerce.  Deep learning uses deep neural network (NN) architectures including densely-connected (DNN), convolutional (CNN), and recurrent (RNN) neural network architectures to perform these tasks \cite{Bishop2006, Lecun2015, Goodfellow2016}. Training a deep neural network can be achieved through supervised optimization in which the network is trained on labeled data set that covers the expected outcomes. Alternatively, an unsupervised approach can be adopted in which optimization is performed on a set of targets without the use of labeled data. 

Application of deep learning in science and engineering can be classified in multiple categories. In the most natural way, DL models are trained on labeled data to classify events and predict system response \cite{Rafiei2017, Rouet-Leduc2017, Perol2018ConvolutionalLocation, Ross2019}. They have been employed to construct surrogate models of the systems for on-the-fly response predictions \cite{Pilania2013, Liang2018, DeVries2018, Wang2018ALearning, Kabacaoglu2019, Bergen2019, Dana2020}. Being trained on a large set of simulated data, they are suitable for real-time inversion and system identification \cite{Wagner2016Theory-guidedScience, Rudy2018}. 

More recently, deep learning has been successfully employed to solve differential and integral equations describing a physical phenomenon without labeled data \cite{Weinan2018,Raissi2019,Kharazmi2019,Fang2019AProblems,Haghighat2020AMechanics,Mao2020Physics-informedFlows,Fuks2020PhysicsMedia}. This class of deep learning, also commonly refereed to as Physics-Informed Neural Networks (PINN), employs neural networks as the approximate solution to the partial differential equations (PDEs). Differentiation is performed by using the Automatic Differentiation (AD) algorithm. The solution, i.e., the parameters of the neural network, is identified by optimizing an objective or loss function that incorporates the PDE residuals, boundary conditions, and initial conditions. PINNs have been used for inverse analysis as well as surrogate modeling in fluid and solid mechanics problems \cite{Raissi2019, Sun2020SurrogateData, Haghighat2020AMechanics,Tartakovsky2020Physics-InformedProblems}. Depending on the application, different variations of PINNs are proposed to improve its accuracy\cite{Meng2019, Haghighat2020AOperatorb, Kharazmi2020, Jagtap2020}. The most attractive feature of PINNs is that it enables forward solution, inversion, surrogate model construction, and data-driven modeling all within the same framework. 

This chapter focuses on the use of deep learning for structural vibration problems through regression techniques, solution methods to the partial differential equations, and inversion techniques.

\section{Deep learning}
\begin{figure}[t]
  \centering
  \includegraphics[width=0.6\textwidth]{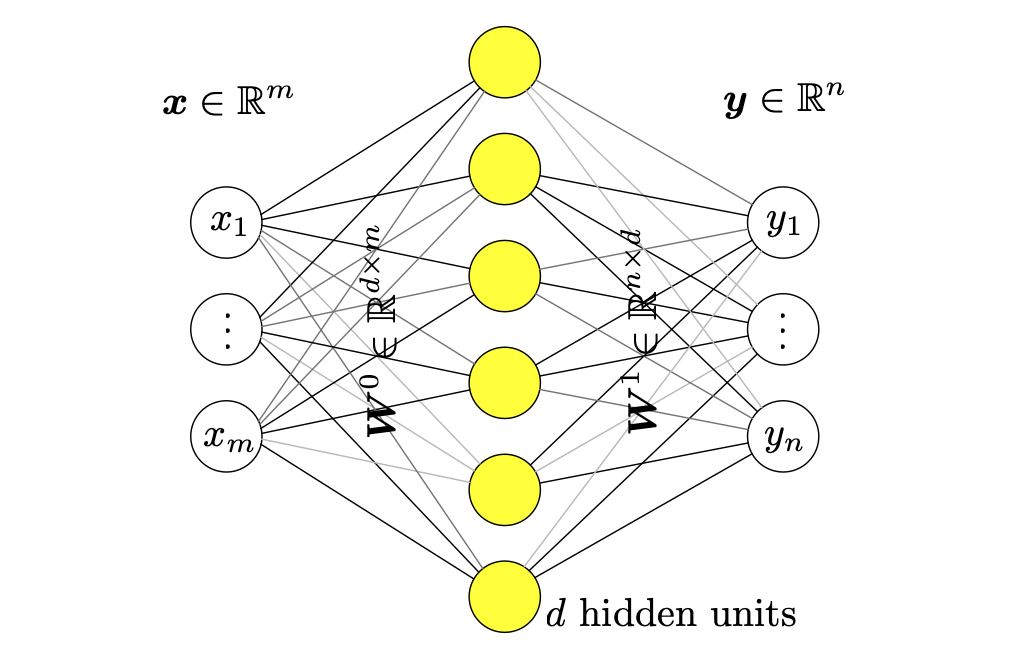}
  \caption{A feed-forward neural network with a single hidden layer with $d$ units, input features $\bs{x} \in \mathbb{R}^m$ and outputs $\bs{y} \in \mathbb{R}^n$. The connections highlight the weights of each layer, i.e. $\bs{W}^0\in\mathbb{R}^{d\times m}$ and $\bs{W}^1 \in \mathbb{R}^{n\times d}$. The bias terms $\bs{b}^0 \in \mathbb{R}^d$ and $\bs{b}^1 \in \mathbb{R} ^n$ sit at each node of this graph. This network constructs the functional form expressed in \cref{eqs:eq1}.}
  \label{fig:NN-struct}
\end{figure}

Deep learning, a subclass of machine learning, employs deep neural networks. A feed-forward neural network with a single hidden layer, and with inputs $\bs{x}\in\mathbb{R}^{m}$, outputs $\bs{y}\in\mathbb{R}^{n}$, and $d$ hidden units is schematically shown in \Cref{fig:NN-struct} and mathematically expressed as 
\begin{align}
  \bs{y} = \bs{W}^1\cdot\sigma(\bs{W}^0\cdot\bs{x} + \bs{b}^0) + \bs{b}^1, \label{eqs:eq1}
\end{align}
where $\sigma$ is an activation function, such as \emph{hyperbolic-tangent}, that makes the transformation nonlinear.  The parameters, $\bs{W}^0\in \mathbb{R}^{d\times m}$ and $\bs{W}^1\in \mathbb{R}^{n\times d}$ are known as \emph{weights} of this transformation and $\bs{b}^0\in \mathbb{R}^{d}$ and $\bs{b}^1\in \mathbb{R}^{n}$ are \emph{biases}. Randomly initialized components of these vectors and matrices define the degrees of freedom (DOF) of this transformation.  Therefore, the transformation \cref{eqs:eq1} has $(d + 1) \times n + (m + 1)\times d$ parameters (DOFs) to be identified. Throughout this study, all of the trainable parameters of the network are collected into a vector of size $D$ and denoted as $\bs{\theta} \in \mathbb{R}^D$. It was shown by \citet{Hornik1989} that this transformation can approximate any measurable function.

To extend the single layer neural network \cref{eqs:eq1} to multi-layers, let us define the transformation $\Sigma$ as $\Sigma^i(\hat{\bs{x}}^i) := \hat{\bs{y}}^i = \sigma^i(\bs{W}^i\cdot\hat{\bs{x}}^i + \bs{b}^i)$, with $\hat{\bs{x}}^i$ as the input and $\hat{\bs{y}}^i$ as the output of any hidden layer $i$, $\bs{x}=\hat{\bs{x}}^0$ as the main input to the network, and $\bs{y}=\Sigma^L(\hat{\bs{x}}^L)$ as the final output of the network. Thus, we can construct a general neural network with $L$ hidden layers as composition of $\Sigma^i$ transformation as
\begin{align}
  \bs{y} = \Sigma^L \circ \Sigma^{L-1} \circ \dots \circ \Sigma^0(\bs{x}). \label{eqs:eq2}
\end{align}
This chapter focuses on regression-type problems or solution methods for PDEs; therefore, we only employ this type of networks, known as \emph{densely-connected} networks or \emph{multilayer perceptron models} (when \emph{sigmoid} activation functions are employed). There are different variations of neural networks that are designed for other applications including \emph{Convolutional} and \emph{Recurrent} networks \cite{Goodfellow2016} for computer vision and natural language processing, respectively. In the context of Physics-Informed Neural Networks, these networks are not currently applicable for solution and inversion of the partial differential equations considered in this chapter. 




The initial values for $\bs{y}(\bs{x}^*)$ are random since the parameters of neural network, i.e. $\bs{\theta}$, are randomly initialized. These parameters are identified using supervised or unsupervised optimization. As an example, the parameters of a regression problem can be identified by optimizing the following objective (loss) function, 
\begin{align}
  \bs{\theta}^* = \argmin_{\bs{\theta} \in \mathbb{R}^D} \mc{L}(\bs{\theta}) := |\bs{y}(\bs{x}^*; \bs{\theta}) - \hat{\bs{y}}^*| \quad \mathrm{where} \quad |\circ| = \mathrm{MSE}(\circ). \label{eqs:eq3}
\end{align}
Here, the asterisked variables $\circ^*$ are the training or collocation points. The loss term is constructed using the mean-squared error (MSE). Optimizing the loss function results in values for the parameters of neural network, i.e., $\bs{\theta}^*$, that minimizes the error between network outputs $\bs{y}(\bs{x}^*_i)$ and their expected values $\bs{y}_i^*$. We will discuss optimization techniques in the following sections.

\section{Physics-Informed Neural Networks}

Physics-Informed Neural Networks (PINN) proposed by Raissi et al. \cite{Raissi2019} is a robust approach to solve differential equations and to perform inversion. According to this architecture, the solution variables are approximated using neural networks with space and time variables as the network features. Differentiation is performed using Automatic Differentiation (AD) \cite{Gune2018}. Initial and boundary conditions as well as differential equations are included in the loss function. The parameters of the neural network are identified by optimizing the loss function. The main advantage of this approach is that the given data related to any variable can be incorporated into the loss function. 

This approach can be illustrated by considering the following Laplace's equation
\begin{align}
 & \kappa \nabla^2 f = \kappa \left(\frac{\partial^2}{\partial x^2} + \frac{\partial^2}{\partial y^2}\right) f = 0, &\mathrm{for}&~ \bs{x} \in \Omega, \label{eqs:eq4}\\
  &f = \bar{f}, &\mathrm{for}&~ \bs{x} \in \Gamma_f, \label{eqs:eq5}\\
  &\kappa \frac{\partial f}{\partial n} = \kappa \nabla f \cdot \bs{n} = \bar{q}, &\mathrm{for}&~ \bs{x} \in \Gamma_q, \label{eqs:eq6}
\end{align} 
where $\kappa$ is a material parameter, $\Omega$ and $\Gamma = \Gamma_f \cup \Gamma_q$ describe the domain and its boundary, respectively. The domain is subjected to Dirichlet and Neumann boundary conditions $\bar{f}$ and $\bar{q}$ on $\Gamma_f$ and $\Gamma_q$, respectively. 

Assuming the solution to this problem is $f = f(x,y)$, the solution can be approximated using a neural network expressed as

\begin{align}
  f(x,y) \approx \hat{f}(x,y; \bs{\theta}): (x,y) \rightarrow \mc{N}_f(x,y; \bs{\theta}). \label{eqs:eq7}
\end{align}

To find the solution to the Laplace's equation, i.e., to find $\bs{\theta}^* \in \mathbb{R}^D$ that satisfies relations \cref{eqs:eq4,eqs:eq5,eqs:eq6}, we need to perform the following optimization: 
\begin{align}
  \begin{split}  
    \bs{\theta}^* = \argmin_{\bs{\theta} \in \mathbb{R}^D} \mc{L} &:= \lambda_1 |\kappa (\frac{\partial^2 \hat{f}}{\partial x^2} + \frac{\partial^2 \hat{f}}{\partial y^2})| \\ 
    &+ \lambda_2 |\delta(\bs{x} \in \Gamma_f)(\hat{f} - \bar{f})| \\ 
    &+ \lambda_3 |\delta(\bs{x} \in \Gamma_q)(\kappa \frac{\partial \hat{f}}{\partial n} - \bar{q})| \label{eqs:eq8}
  \end{split}
\end{align}
where $\delta$ is the Dirac function and $|\circ|$ represents the MSE norm. $\lambda_i$ are the weights associated with each term in the loss function. Optimization of the total loss $\mc{L}$ results in the solution to the Laplace equation \cref{eqs:eq4,eqs:eq5,eqs:eq6}. 

The inversion with PINN can be illustrated by considering the same problem. In the inversion process, we will find the parameter of the system, i.e., $\kappa$ based on the available measurements. Let us assume that the field variable $f$ is measured on a set of discrete points $\bs{x}^* \in \Omega$. To use PINN for inversion, we need to also define $\kappa$ as a trainable parameter. Additionally, we can incorporate a loss term for the measurements on $f$. Therefore, the optimization problem \cref{eqs:eq8} is modified as follows: 
\begin{align}
  \begin{split}
    \kappa^*, \bs{\theta}^* = \argmin_{\kappa\in\mathbb{R}, ~\bs{\theta} \in \mathbb{R}^D} \mc{L} &:= \lambda_1|\kappa (\frac{\partial^2 \hat{f}}{\partial x^2} + \frac{\partial^2 \hat{f}}{\partial y^2})| \\ 
    &+ \lambda_2|\delta(\bs{x} \in \Gamma_f)(\hat{f} - \bar{f})| \\
    &+ \lambda_3|\delta(\bs{x} \in \Gamma_q)(\kappa \frac{\partial \hat{f}}{\partial n} - \bar{q})| \\
    &+ \lambda_4|\delta(\bs{x}^*\in\Omega)(\hat{f} - f^*)|.\label{eqs:eq9}
  \end{split}
\end{align}
where $f^*$ are measurements at $\bs{x}^*$ points. There exist other variations of PINNs to address different types of PDEs including Variations, Parareal, Domain Decomposed, and Nonlocal PINNS \cite{Kharazmi2019, Meng2019, Jagtap2020, Haghighat2020AOperatorb}. However, we limit the scope of this chapter to the standard PINN.

\section{Training Neural Networks}
The goal of training or optimization is to find a set of parameters $\bs{\theta}^*$ that minimizes the loss function $\mc{L}(\bs{\theta}): \mathbb{R}^D \rightarrow \mathbb{R}$, where $D$ is the total number of parameters. For the purpose of illustration, considering the loss function associated with the Laplace's equation, Taylor's theorem for the loss function $\mc{L}$ leads to  
\begin{align}
  \mc{L}(\bs{\theta} + \bs{\vartheta}) = \mc{L}(\bs{\theta}) + \nabla \mc{L}(\bs{\theta})\cdot \bs{\vartheta} + \frac{1}{2} \bs{\vartheta}\cdot\nabla^2\mc{L}(\bs{\theta} + t \bs{\vartheta})\cdot \bs{\vartheta}, \label{eqs:eq10}
\end{align}
for $t \in (0,1)$, where $\nabla$ and $\nabla^2$ are gradient and \emph{Hessian} operators. The necessary conditions for $\bs{\theta}^*$ to be a local minimizer is that the gradient vector $\nabla \mc{L}(\bs{\theta}^*) = \bs{0}$ and the Hessian matrix to be positive definite, i.e., $\bs{v}\cdot\nabla^2\mc{L}(\bs{\theta}^*)\cdot\bs{v}>0$ for any arbitrary $\bs{v}\in\mathbb{R}^D$. These conditions imply that when $\mc{L}$ is convex, any local minimizer $\bs{\theta}^*$ is a global minimizer. Note that for neural networks, the loss function is often non-convex. 

There are two family of algorithms to solve this optimization problem.  The first one is known as \emph{line search} or first-order methods and the second one as the \emph{trust region} or second-order methods \cite{Robinson2006}. Line search methods are based on moving along a direction that most rapidly decreases the loss function. It can be easily shown that if we write the Taylor's expansion for $\mc{L}(\bs{\theta} + \alpha \bs{\vartheta})$ up to the first order, the maximum drop in $\mc{L}(\bs{\theta})$ happens when we move in the $\bs{\vartheta} = -\nabla \mc{L}$ direction. This builds the foundation for the first-order \emph{gradient descent} methods. There are numerous ways to choose a \emph{step-size} or \emph{learning-rate} $\alpha$. Among those, the most commonly used approach is \emph{Adam} methods, where $\alpha$ is chosen adaptively for each parameter $\theta_i$ based on the first and second momentum updates \cite{Kingma2015Adam:Optimization}. 

The most fundamental second-order method also known as \emph{trust-region} is the Newton's method, which can be derived by setting the variation of the second order Taylor's expansion of the loss $\mc{L}(\bs{\theta} + \bs{\vartheta})$ to zero. This results into a direction update $\bs{\vartheta} = -(\nabla^2 \mc{L})^{-1} \cdot \nabla \mc{L}$. By comparison to the first order updates, the learning rate $\alpha$ is replaced by the inverse of the Hessian matrix. Evaluation of the inverse of Hessian matrix and its memory storage can become quickly unmanageable for high-dimensional optimization problems. Therefore, approximations of the inverse of Hessian matrix is desirable. There are numerous methods that are designed to address this drawback. Among those, BFGS algorithm and its variations are among the most commonly known approaches.  We refer the interested readers to \cite{Robinson2006, Sun2019OptimizationAlgorithms} for additional details. 

As we find, relations \cref{eqs:eq8} and \cref{eqs:eq9} represent multi-objective optimization and $\lambda_i$ highlight the scaling factor for each term. These factors are used to scale each term in the loss function so that the optimizer can find the global minimum more efficiently. While they can be tuned manually to increase or decrease the importance of each term, there are numerous suggestions that can be used to automatically scale each term in the loss function, including a weighted sum approach based on the value of loss function \cite{Kim2005AdaptiveGeneration}. For the physics-informed neural networks, \citet{Wang2020UnderstandingNetworks} propose a scaling approach based on balancing the distribution of gradients of each term in the loss function. This approach is adopted for the problems solved in this chapter.

\section{Applications of Deep Learning for Data Representation}

This section presents the applications of deep learning by considering vibration problems relevant to civil engineering. These applications concern regression and smoothing data, solution and characterization of single-degree of freedom rigid-block, membranes, and plates. All the problems presented in this chapter are solved using the SciANN \cite{Haghighat2019SciANN:Networks} project and the codes are shared in this github repository \href{https://github.com/sciann/sciann-applications/SciANN-Vibrations}{SciANN-Vibrations}.
\subsection{Polynomial regression}

Polynomial regression can be performed through a neural network as
\begin{align}
  \bs{y} = \bs{b} + \bs{W}_1\cdot\bs{x} + \bs{W}_2\cdot\bs{x}^2 + \dots + \bs{W}_p\cdot\bs{x}^p
\end{align}
This defines a polynomial regression model of order $p$. The main advantage of constructing such a model using neural networks is that we can leverage the AD algorithm for optimization as well as for the PINN framework. The input features are denoted as $(\bs{x}, \bs{x}^2, \dots, \bs{x}^p)$. For $p=1$, it a linear regression model and $p=2$ is a quadratic regression model. 

The simplest neural network is a linear model expressed as
\begin{align}
  \bs{y} = \mathbf{W}\cdot\bs{x} + \mathbf{b}
\end{align}
The dimension of weight matrix $\bs{W}$ and bias vector $\bs{b}$ depend on the size of inputs $\bs{x}$ and outputs $\bs{y}$. 

\begin{figure}[t]
  \centering
  \includegraphics[width=1.0\textwidth]{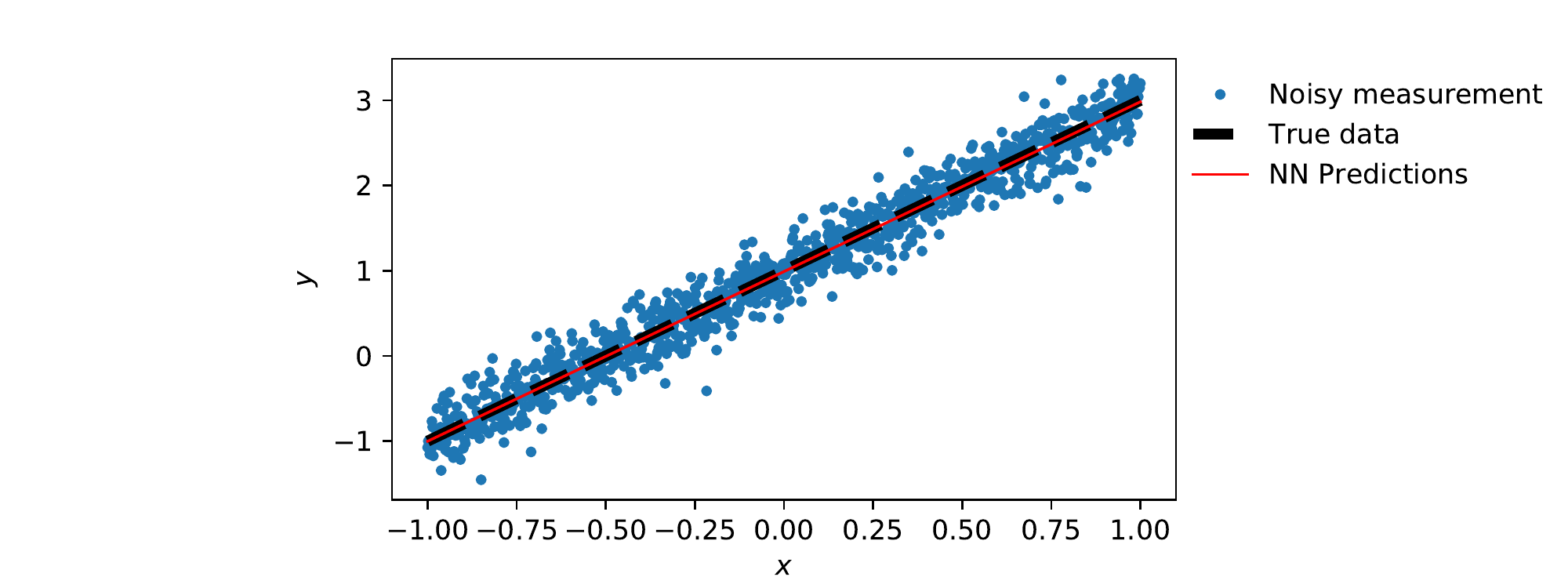}
  \caption{Use of a simple neural network architecture to perform linear regression on noisy measurements. The optimizer correctly approximates the \emph{true} function. }
  \label{fig:linear-regression}
\end{figure}

As shown in \Cref{fig:linear-regression}, for a given noisy dataset $\mathbf{X},\mathbf{Y}^*$ containing $N$ data points, the input and output have a dimension of 1, i.e., $\bs{W}\in\mathbb{R}^1$ and $\bs{b}\in\mathbb{R}^1$. Therefore, the total number of parameters for the training is 2, i.e., $\bs{\theta} = (\bs{W}, \bs{b}) \in\mathbb{R}^2$. According to \cref{eqs:eq3}, the optimization problem can be defined as 
\begin{align}
  \argmin_{\bs{\theta}\in\mathbb{R}^2} \mc{L} := |y(x_i) - y^*_i| =  \sum_{i=1}^N  \frac{\Big((W~x_i + b) - y^*_i\Big)^2}{N} 
\end{align}
Minimization of this problem results in optimized values for $W\approx 2.0$ and $b\approx 1.0$. 

\begin{figure}[t]
  \centering
  \includegraphics[width=1.0\textwidth]{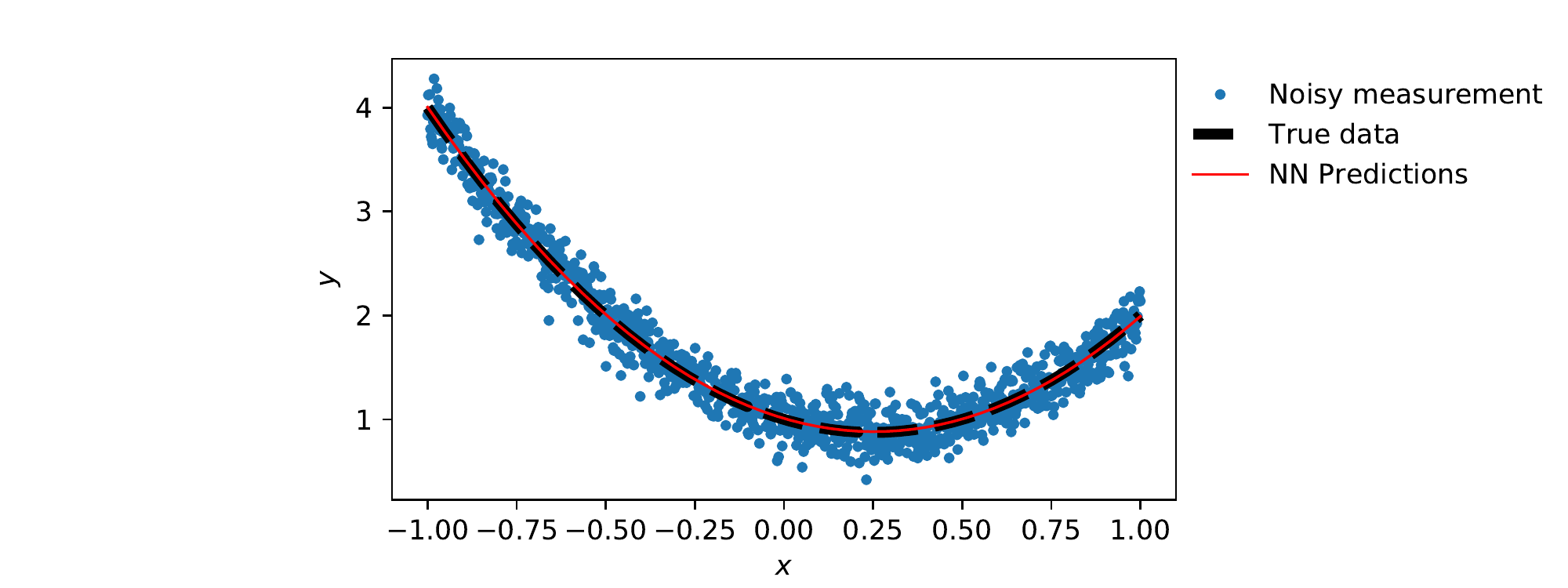}
  \caption{Use of a simple neural network for quadratic regression on the noisy data. The inputs features are $x, x^2$. The network correctly identifies the \emph{true} underlying model for data.}
  \label{fig:quadratic-regression}
\end{figure}

For the noisy dataset $(\mathbf{X}, \mathbf{Y}^*)$ shown in  \Cref{fig:quadratic-regression}, we can construct a quadratic regression model with neural networks and train the model on this dataset. Accordingly, the total number of parameters for this model is 3, i.e., $\bs{\theta} \in \mathbb{R}^3$. The optimization problem becomes:
\begin{align}
  \argmin_{\bs{\theta}\in\mathbb{R}^3} \mc{L} := |y(x_i) - y^*_i| =  \sum_{i=1}^N  \frac{\Big((W_1~x_i + W_2~x_i^2 + b) - y^*_i\Big)^2}{N} 
\end{align}
Minimization of this objective functions results in optimized values for $W_1\approx -1$, $W_2\approx 2$ and $b\approx 1$.

\subsection{Smoothing noisy vibration measurements}

\begin{figure}[t]
  \centering
  \includegraphics[width=1.0\textwidth]{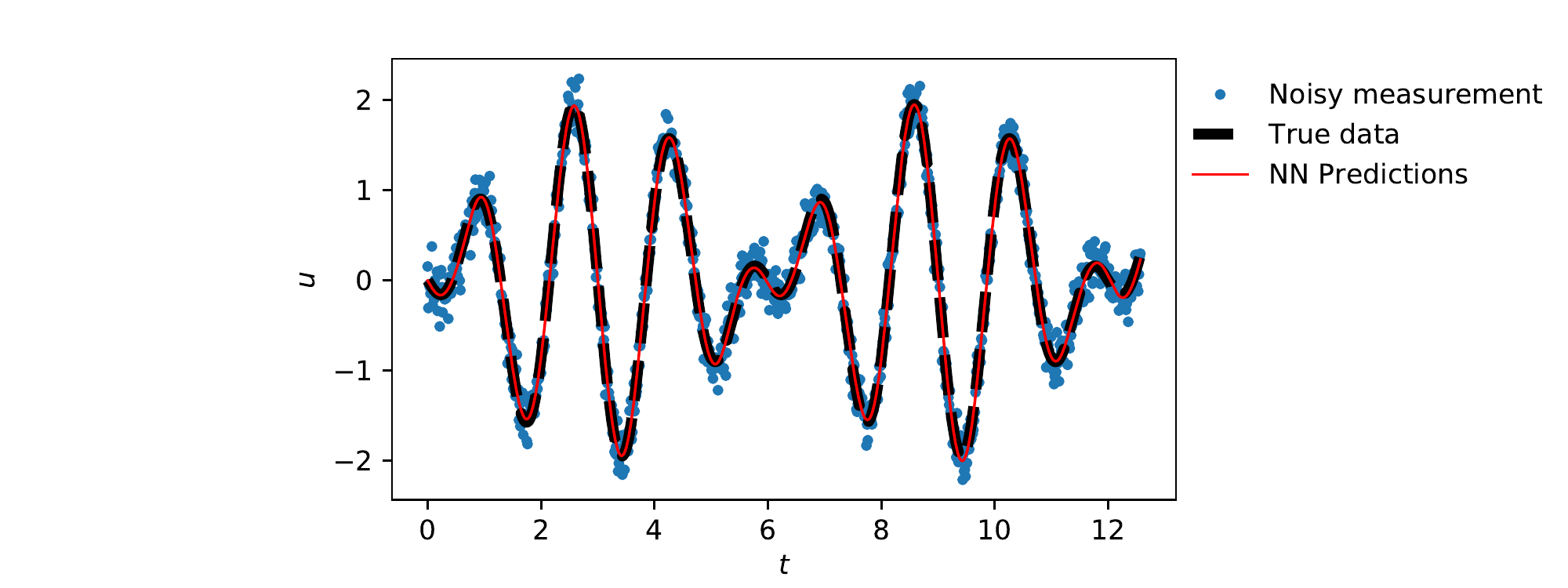}
  \caption{Use of a single-layer network for regression on noisy periodic data. The network accurately predicts the underlying model of the data.}
  \label{fig:rnn-regression}
\end{figure}

Noisy vibration measurements are considered to illustrate a more complex neural network architecture for deep learning. The data is generated by employing the following periodic function, also shown in \Cref{fig:rnn-regression}:
\begin{align}
  y = A~(\sin \omega t - \beta \sin \bar{\omega} t) + \varepsilon, \quad \varepsilon \sim N(0, \sigma^2)
\end{align}
with $A=1$, $\beta = 1$, $\omega=2\pi/2$, and $\bar{\omega} = 2\pi/1.5$. 

The objective is to construct a smooth and continuous representation of the noisy measurements. This representation is often necessary to evaluate the parameters of the system.  Since the data set is intrinsically periodic, we set up a Fourier network with a sinusoidal activation function to perform the regression. The functional form of the neural network is expressed as 
\begin{align}
  \bs{y} = \bs{W}^2 \cdot \sin(\bs{W}^1 \cdot \bs{x} + \bs{b}^1) + \bs{b}^2.
\end{align}
This single layer neural network represents the following Fourier series
\begin{align}
  y = \sum_{i=1}^{N_n} W^2_i \sin(W^1_i t + b^1_i) + b^2_i,
\end{align}
where $N_n$ is the number of neurons in the hidden layer. We determine different frequencies $\bs{W}^1$, phases $\bs{b}^1$, and amplitudes  $\bs{W}^2$. As shown in \Cref{fig:rnn-regression}, this network identifies the true distribution of the data without \emph{overfitting} to the noise.

\section{Deep Learning for Solution and Inversion of Vibration Problems}
This section presents the application of deep learning for solving and inversion of time dependent ordinary and partial differential equations. 

\subsection{Forced vibration spring-mass problem}
The forced vibration of spring-mass system can be described by
\begin{align}
  \ddot{u} + \omega^2~u &= f_0 \sin \bar{\omega}t, \\
  u(t=0) &= 0, \\
  \dot{u}(t=0) &= 0,
\end{align}
where $\omega^2 = k/m$ is the natural frequency of the spring-mass system with $m$ and $k$ as its mass and stiffness, respectively.  With $F_0 = m~f_0$ and $\bar{\omega}$ as the amplitude and frequency of the applied external load, the analytical solution to this system is expressed as
\begin{align}
  u(t) = \frac{F_0}{\omega^2 - \bar{\omega}^2} \Big( \sin \bar{\omega} t - \beta \sin \omega t\Big),
\end{align}
with $\beta$ as $\beta = \bar{\omega}/\omega$. 

Neural networks can be used to solve the response of this initial value problem.  Employing the PINN framework explained earlier, the solution space can be approximated with a multi-layer neural network $\hat{u} = \mathcal{N}_u(t; \bs{\theta})$, and the ODE residual is evaluated using AD. The optimization problem is then expressed as:
\begin{align}
  \argmin_{\bs{\theta} \in \mathbb{R}^D} \mathcal{L}(\bs{\theta}) := \lambda_1|\frac{\partial^2 \hat{u}}{\partial t^2} + \omega^2\hat{u} - f_0 \sin \bar{\omega}t | + \lambda_2 |\hat{u}(t=0)| + \lambda_3|\frac{\partial\hat{u}}{\partial t}(t=0)|
\end{align}

\begin{figure}[t]
  \centering 
  \includegraphics[width=0.9\textwidth]{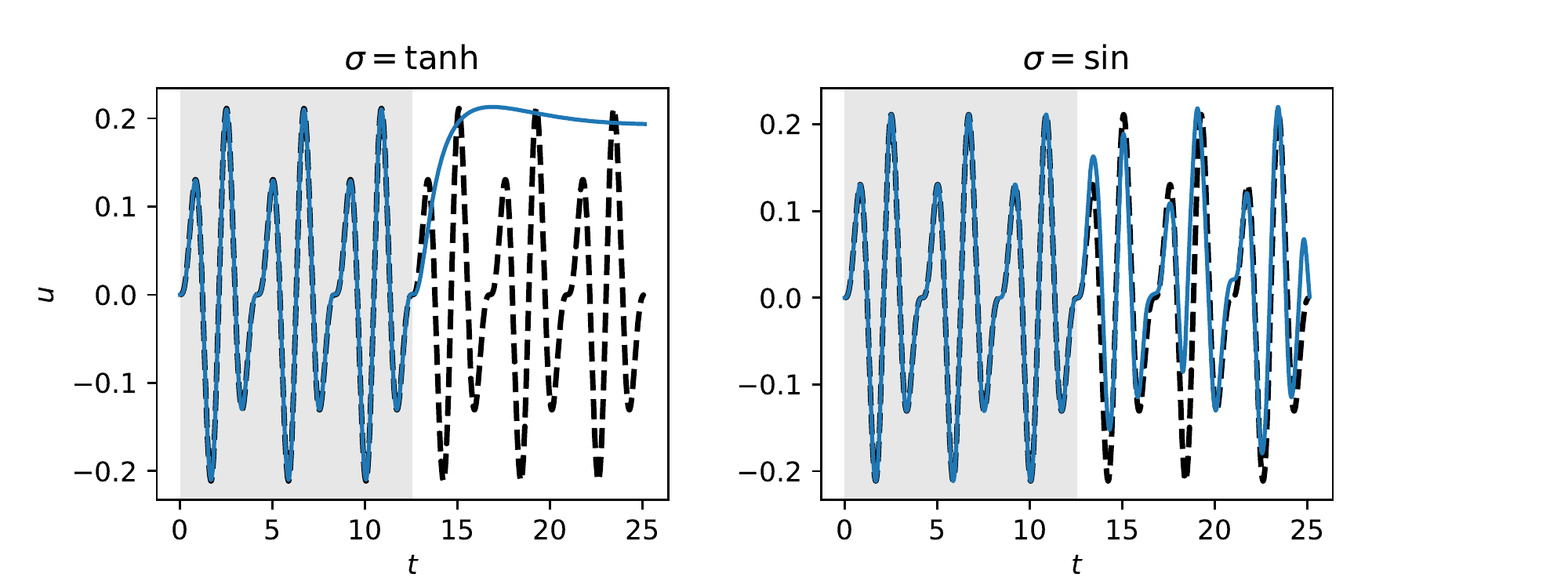}
  \caption{Forward solution of the forced vibration problem for $\omega=3$, $\beta=1.5$, and $F_0=1$. Neural network has 4 hidden layers with 20 neurons. For the left plot, $\mathrm{tanh}$ activation function is used while for the right plot, the activation function is $\mathrm{sin}$. The highlighted grey area show the training domain. The network on the right provides better predictive capability and therefore has learned the underlying dynamics imposed on the network. }
  \label{fig:forced-vibration}
\end{figure}

\Cref{fig:forced-vibration} presents the results for a system with $\omega=3$, $\beta=1.5$, and $F_0=1$. The displacement $u$ is approximated with a network with 4 hidden layers, each with 20 neurons, and with hyperbolic-tangent as the activation function.  The data is generated for $t\in [0, 4\pi]$ and tested for $t\in [0, 8\pi]$.  Two networks with hyperbolic-tangent and sinusoidal activation functions are considered, both with 4 hidden layers and 20 neurons per layer.  The results show much more accuracy and predictive capability for the sinusoidal network, as it is naturally a more suitable choice due the periodic nature of the data.

\begin{figure}[t]
  \centering 
  \includegraphics[width=0.9\textwidth]{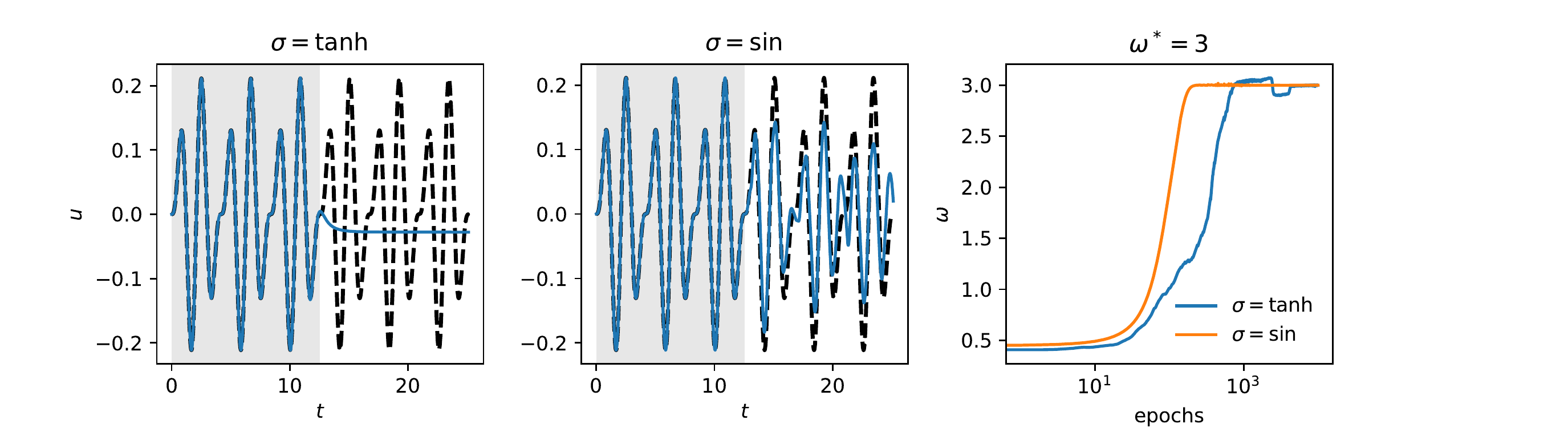}
  \caption{Inversion solution of the forced vibration problem on a dataset generated with $\omega=3$, $\beta=1.5$, and $F_0=1$. Neural network has 4 hidden layers with 20 neurons. For the left plot, $\mathrm{tanh}$ activation function is used while for the right plot, the activation function is $\mathrm{sin}$. The highlighted grey area show the training domain. The network on the right provides better predictive capability and therefore has learned the underlying dynamics imposed on the network. }
  \label{fig:forced-vibration-inversion}
\end{figure}

Next, we assume that a displacement history for the problem is given, and we want to characterize the system, i.e., identifying the natural frequency of the problem. Mathematically, the optimization problem is now expressed as 
\begin{align}
  \argmin_{\bs{\theta}\in\mathbb{R}^D,~\omega\in\mathbb{R}} \mc{L}(\bs{\theta}, \omega) := \lambda_1 |\frac{\partial^2 \hat{u}}{\partial t^2} + \omega^2 \hat{u}^2 - f_0 \sin \bar{\omega}t| + \lambda_2| \hat{u} - u^* |.
\end{align}
This framework can be easily adopted for both forward and inversion problems. The result of this optimization is a neural network model that gives the solution for the discrete displacement data as well as identification of the frequency of the system. The results are shown in \cref{fig:forced-vibration-inversion}. 

\subsection{Free vibration of rectangular membrane}

A rectangular membrane with unit dimensions $1\times 1$ is excited by an initial displacement $u = \sin \pi x \sin \pi y$. The governing PDE and boundary conditions are expressed as 
\begin{align}
  &c \nabla^2 u = c \left(\frac{\partial^2}{\partial x^2} + \frac{\partial^2}{\partial y^2}\right) u = \frac{\partial^2 u}{\partial t^2}, &\mathrm{for}&~ \bs{x} \in [0, 1]^2, ~t>0, \\
  &u = 0, &\mathrm{for}&~ \bs{x} \in \Gamma_u, \\
  &u = \sin{\pi x}\sin{\pi y}, &\mathrm{for}&~ t = 0, \\
  &\frac{\partial u}{\partial t} = 0, &\mathrm{for}&~ t = 0,
\end{align}
where $u$ is the displacement, and $c$ is the wave propagation speed. The analytical solution to governing equation is $u(x, y, t) = \sin \pi x \sin\pi y \cos\sqrt{2} \pi t$. 

\begin{figure}[t]
    \centering
    \includegraphics[width=1.0\textwidth]{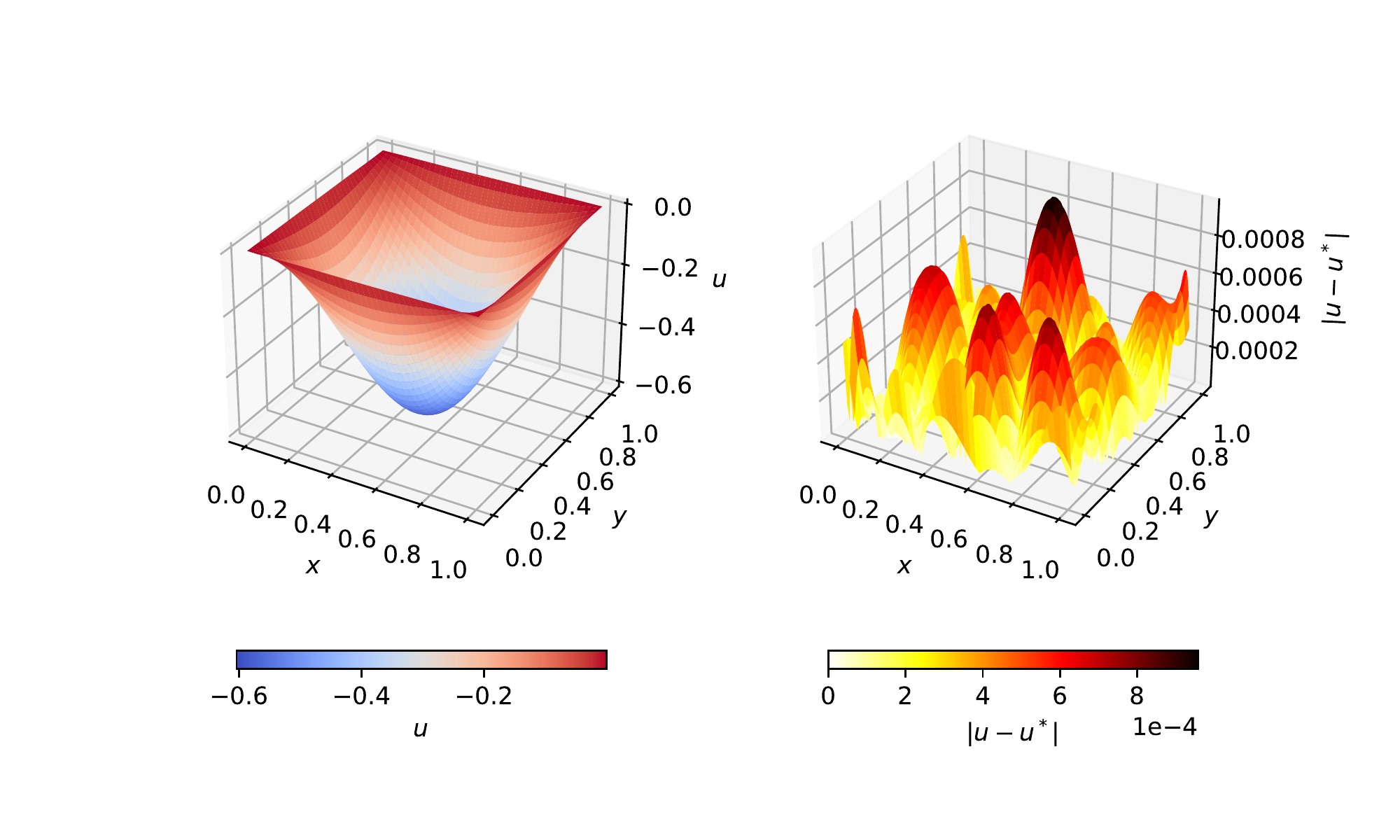}
    \caption{The results of the PINN solution to the membrane problem at $t=0.5$. (Left) out of plane deflection of the membrane. (Right) the absolute error between network predictions and analytical solution. }
    \label{fig:membrane-forward}
\end{figure}

To solve this problem using PINN, the solution space can be constructed using a neural network $\hat{u}(x,y, t; \bs{\theta})=\mc{N}_u(x,y, t; \bs{\theta})$. The forward optimization problem is expressed as
\begin{align}
  \begin{split}  
    \argmin_{\bs{\theta} \in \mathbb{R}^D} \mc{L}(\bs{\theta}) &:= \lambda_1|c (\frac{\partial^2 \hat{u}}{\partial x^2} + \frac{\partial^2 \hat{u}}{\partial y^2}) - \frac{\partial^2 \hat{u}}{\partial t^2}| \\ 
    &+ \lambda_2|\delta(\bs{x} \in \Gamma_u)(\hat{u})| \\ 
    &+ \lambda_3|\delta(t)(\hat{u} - \sin{\pi x}\sin{\pi y})|\\ 
    &+ \lambda_4|\delta(t)(\frac{\partial \hat{u}}{\partial t})|. 
  \end{split}
\end{align}
To perform the optimization, a collocation grid is constructed with $N_x = N_y = 40$, $N_t =  20$, and terminal time is $T=1/(2\sqrt{2})$ unit. The stiffness coefficient $c$ is specified as $c=1$. We use a neural network with 4 hidden layers, each containing 20 neurons, and with sinusoidal activation function. The training is performed by using $20,000$ epochs. The out of plane displacement as well as the error plot are shown in \cref{fig:membrane-forward}. As we find, the PINN framework can accurately solve this problem. 

\begin{figure}[t]
  \centering 
  \includegraphics[width=1.0\textwidth]{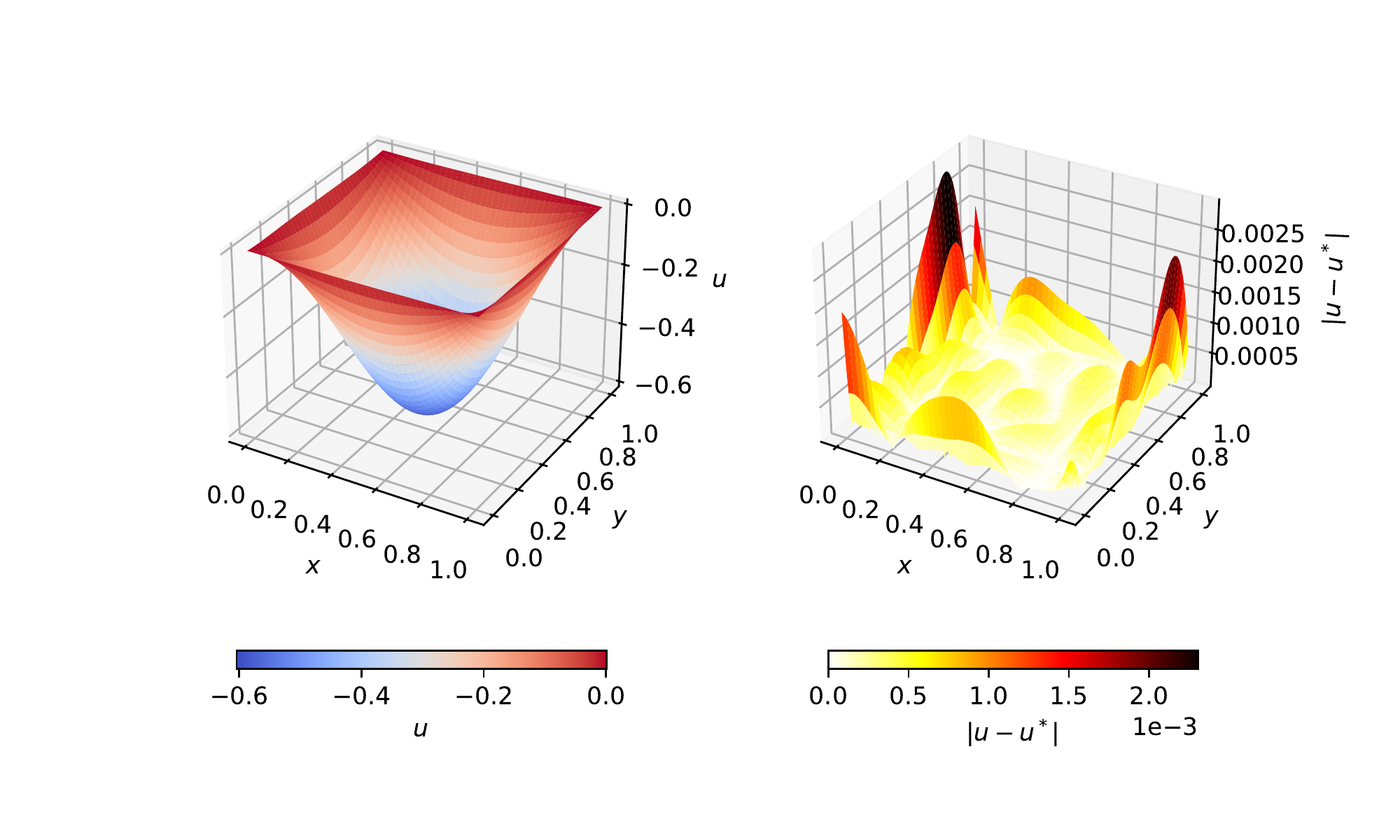}
  \includegraphics[width=0.4\textwidth]{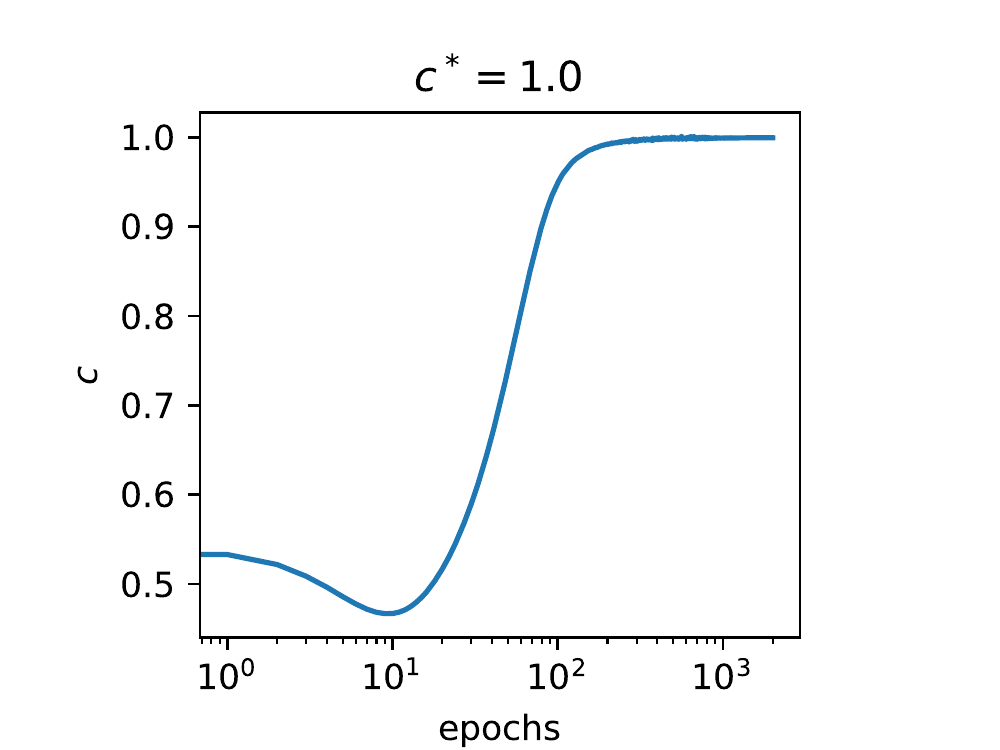}
  \caption{Inversion solution to the membrane problem. (Top-left) The out of plane deformation of the membrane at $t=0.5$. (Top right) The absolute error at $t=0.5$ for the out of plane deformation. (Bottom) The training history for the system parameter $c$, which has been accurately predicted by the PINN framework. }
  \label{fig:membrane-inversion}
\end{figure}

For the inverse problem, the dataset for training is generated using the analytical solution. Same test case is used but with a terminal time of $T=1.0s$ with discretization parameters of $N_x = N_y = N_t = 40$. The optimization problem becomes
\begin{align}
  \argmin_{c\in\mathbb{R}, ~\bs{\theta} \in \mathbb{R}^D} \mc{L}(c, \boldsymbol{\theta}) &:= \lambda_1|c (\frac{\partial^2 \hat{u}}{\partial x^2} + \frac{\partial^2 \hat{u}}{\partial y^2}) - \frac{\partial^2 \hat{u}}{\partial t^2}| + \lambda_2|\hat{u} - u^*|
\end{align}
This optimization results in a value for parameter $c$ as well as a solution, in the form of a neural network, that satisfies the membrane governing equation. The results are shown in \Cref{fig:membrane-inversion}. As you find, the PINN framework accurately captures the parameter of the problem, i.e., $c$, as well as the solution to this problem. 

\subsection{Free vibration of a rectangular plate}
The governing equation describing the out of plane motion of a plate is described as 

\begin{align}
  D \nabla^4 u &= D \left(\frac{\partial^4}{\partial x^4} + 2\frac{\partial^4}{\partial x^2y^2} + \frac{\partial^4}{\partial y^4}\right) u = -\rho \frac{\partial^2 u}{\partial t^2},
\end{align}

where $D$ is the flexural stiffness given as $D = \frac{Eh^3}{12(1-\nu^2)}$, with $E$ as Young's modulus, $h$ as thickness of the plate, and $\nu$ as the Poisson's ratio. The in-plane dimensions of the plate are specified as unity. The plate is subjected to the following initial and boundary conditions:
\begin{align}
&u = 0,  &\mathrm{for}& \quad \bs{x} \in \Gamma_u, \\
  &\frac{\partial^2 u}{\partial n^2} = 0,  &\mathrm{for}& \quad \bs{x} \in \Gamma_u, \\ 
&u = \sin{\pi x}\sin{\pi y},  &\mathrm{for}&  \quad t = 0, \\
&\frac{\partial u}{\partial t} = 0,  &\mathrm{for}& \quad t = 0.
\end{align}
The analytical solution to the governing equation can be constructed as 
\begin{align}
  u(x, y, t) = \sin (\pi x) \sin (\pi y) \cos (\frac{4 \pi^4 D}{\rho} t). 
\end{align}

\begin{figure}[t]
    \centering
    \includegraphics[width=\textwidth]{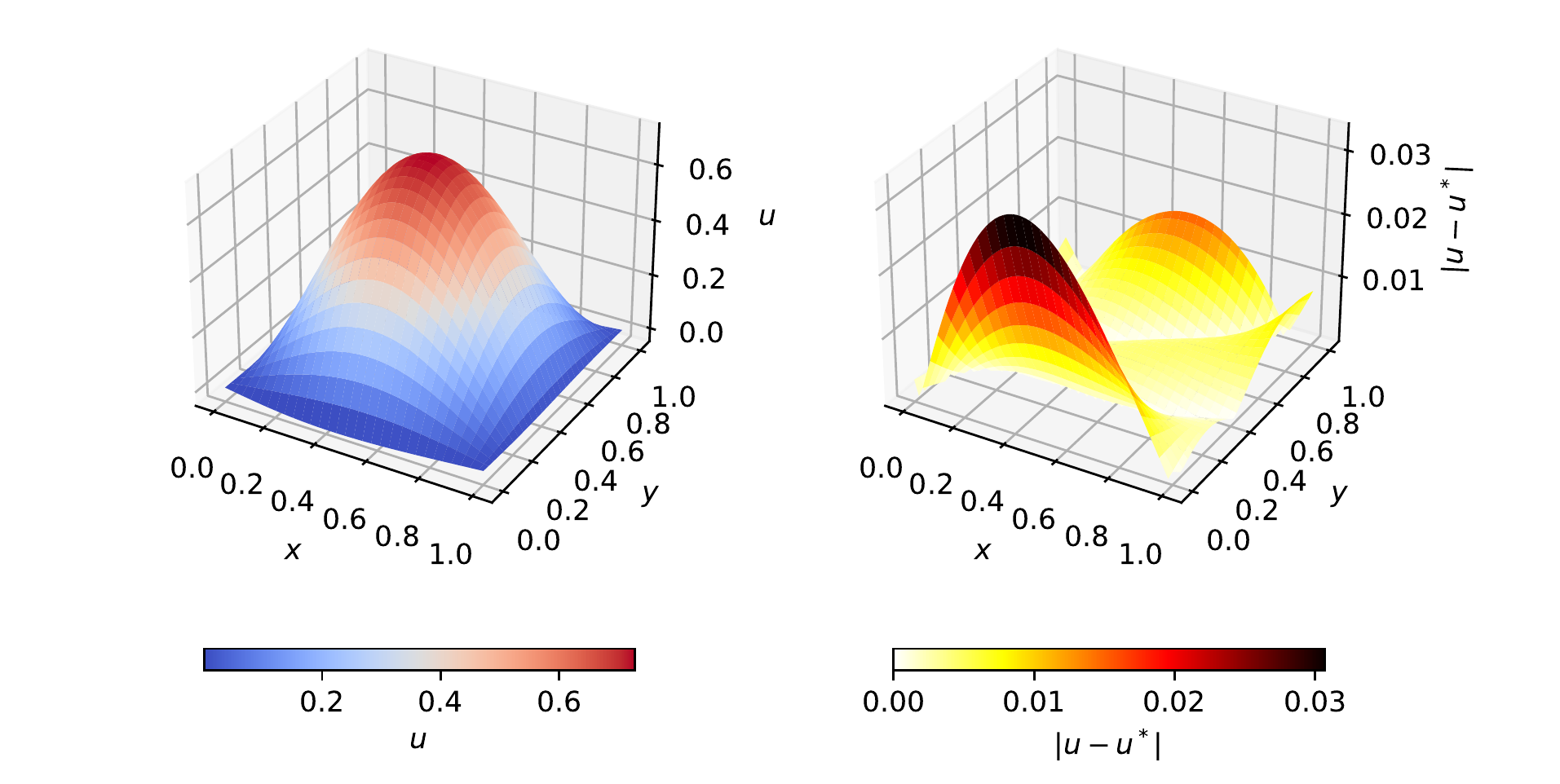}
    \caption{(Left) Predicted out of plane deformation using PINN. (Right) The absolute error between the predicted PINN results and the analytical solution. }%
    \label{fig:plate-forward}%
\end{figure}

The solution to this problem with PINN requires the construction of a neural network as $\hat{u}(x, y, t) = \mathcal{N}_u(x,y,t; \bs{\theta})$. The optimization problem becomes 
\begin{align}
  \begin{split}  
    \argmin_{\bs{\theta} \in \mathbb{R}^D} \mc{L}(\bs{\theta}) &:= \lambda_1|D (\frac{\partial^4 \hat{u}}{\partial x^4} + \frac{\partial^4 \hat{u}}{\partial x^2y^2} + \frac{\partial^4 \hat{u}}{\partial y^4}) + \rho \frac{\partial^2 \hat{u}}{\partial t^2}| \\ 
    &+ \lambda_2|\delta(\bs{x} \in \Gamma_u)(\hat{u})| \\
    &+ \lambda_3|\delta(\bs{x} \in \Gamma_u)(\frac{\partial^2 \hat{u}}{\partial n^2})| \\ 
    &+ \lambda_4|\delta(t)(\hat{u} - \sin{\pi x}\sin{\pi y} / 100)|\\ 
    &+ \lambda_5|\delta(t)(\frac{\partial \hat{u}}{\partial t})|
  \end{split}
\end{align}
The specified plate parameters are: thickness as $h=0.004m$, Poisson's ratio as $\nu=0.25$, Young's modulus as $E=7E+10 Pa$, and density as $\rho=2700kg/m^3$. The optimization is performed with a uniform grid $N_x = N_y = 20$, $N_t =  40$ and a terminal time $T=0.1s$. The network consists of 4 hidden layers with 40 neurons in each layer and sine activation function. The training is conducted with $1000$ epochs and a batch size of $512$. The results are shown in \Cref{fig:plate-forward}. The out of plane displacements predictions, in this case, is less accurate than the membrane problem. This is associated with the higher order of the governing equations and could be improved increasing the weight associated with the boundary conditions. 

\begin{figure}[t]
  \centering 
  \includegraphics[width=1.0\textwidth]{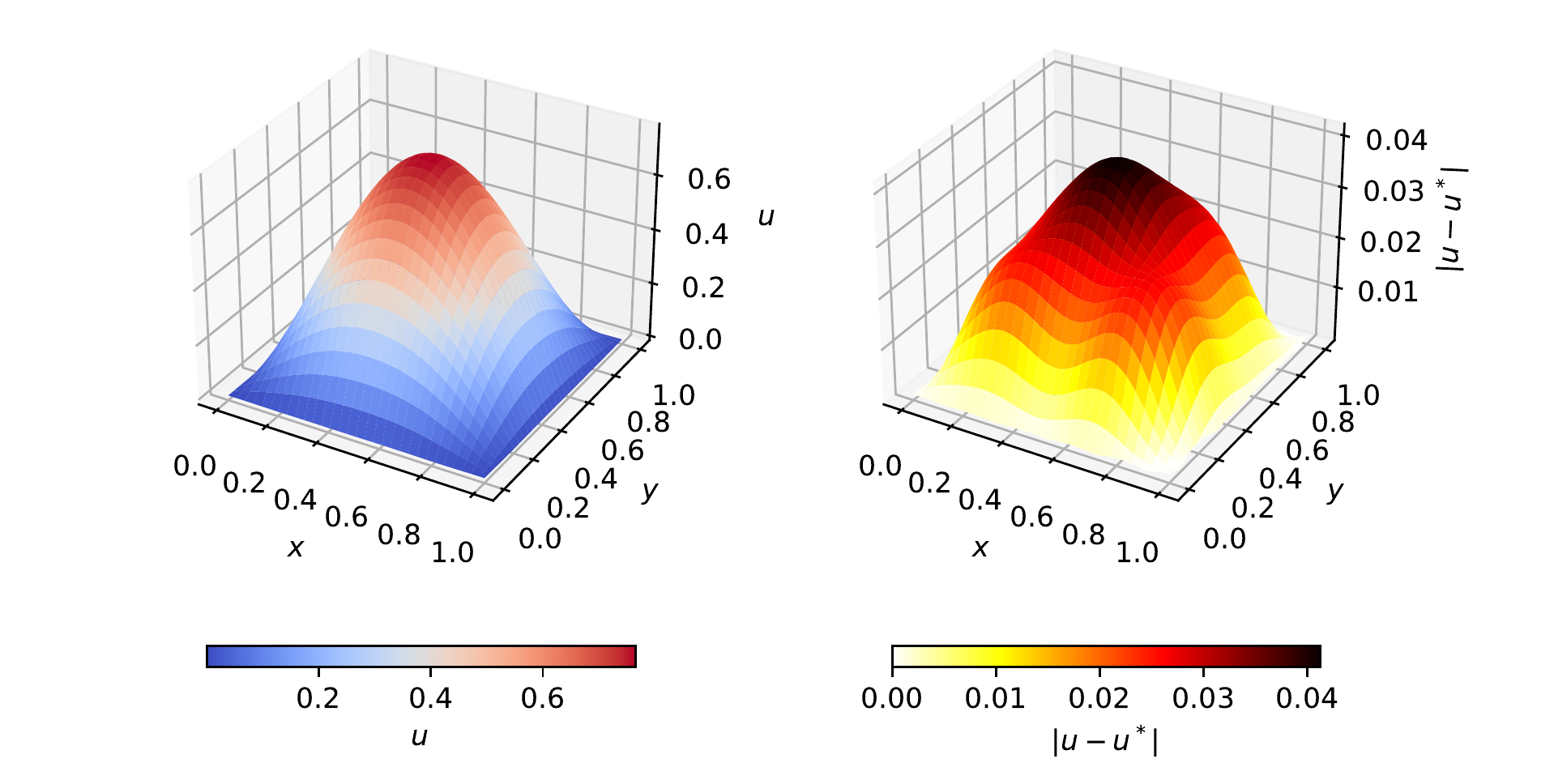}
  \includegraphics[width=0.45\textwidth]{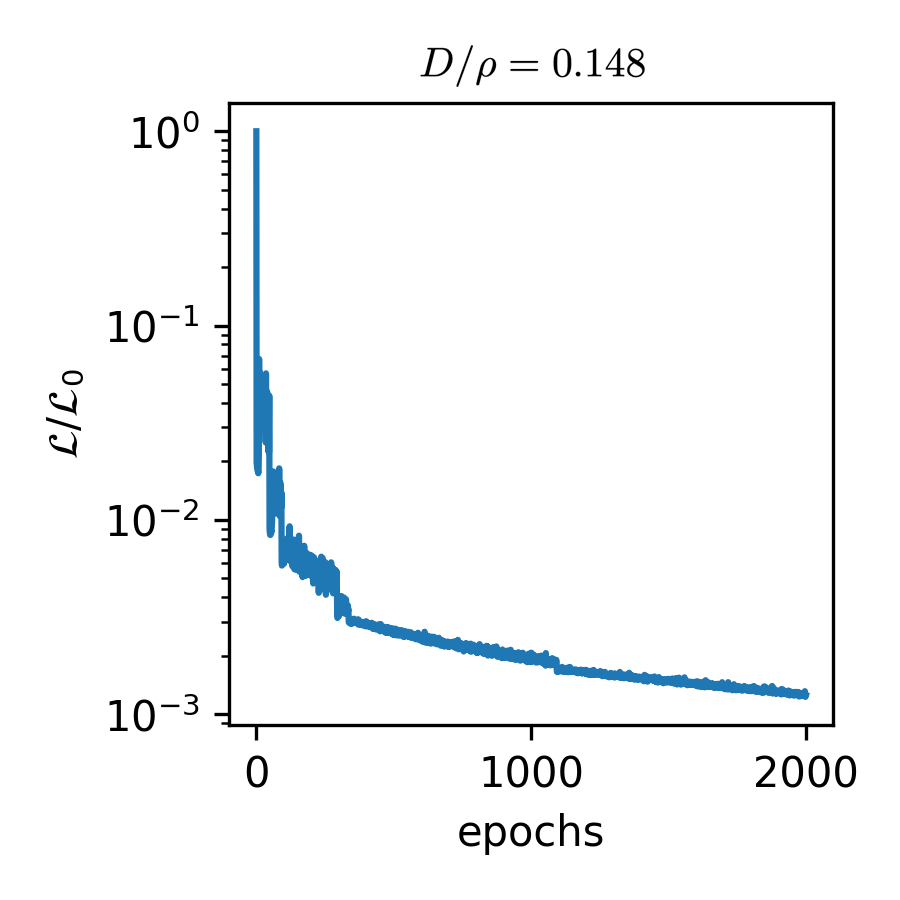}
  \caption{Identification of plate-vibration problem. (Top-left) The solution to the problem using PINN as a result of combined solution and inversion algorithm. (Top-right) The absolute error between neural network predictions and analytical solution for out of plane deformation. (Bottom) Evolution of mean-squared error for the inverse problem with accurate prediction for the value of the normalized parameter $\hat{D}=D/\rho$.}
  \label{fig:plate-inversion}
\end{figure}

For the inversion, the dataset for training is generated using the analytical solution. The same test case is used but with a terminal time of $T=1.0s$ and discretization parameters of $N_x = N_y = N_t = 40$. The optimization problem then becomes 
\begin{align}
  \begin{split}  
    \argmin_{D\in\mathbb{R}, ~\bs{\theta} \in \mathbb{R}^D} \mc{L}(D, \bs{\theta}) &:= \lambda_1|D (\frac{\partial^4 \hat{u}}{\partial x^4} + \frac{\partial^4 \hat{u}}{\partial x^2y^2} + \frac{\partial^4 \hat{u}}{\partial y^4}) + \rho \frac{\partial^2 \hat{u}}{\partial t^2}| \\ 
    &+ \lambda_2|\hat{u} - u^*| 
  \end{split}
\end{align}
Also, the parameter $D$ is normalized using density value as $\hat{D}=D/\rho$ .  The ground truth value of the normalized parameter is $0.147$. The change in loss through the optimization process and inverted parameter $c$ is depicted in \Cref{fig:plate-inversion}.

\section{Discussions and final remarks}
This chapter presents a brief description of deep learning methods and their application concerning structural vibrations.  In describing the applicability of these method, the datasets are generated using the analytical solutions.  However, these DL methods can handle data-driven solutions and data-driven discovery of the parameters.  The results presented in this chapter demonstrate that PINN can solve several types of structural vibration problems with high accuracy. Thus, DL predictions could provide a way to assist in the prediction and optimization of vibration characteristics and system properties. It is worth emphasizing that training neural networks are often challenging due to initialization method and optimization algorithms. Once the correct network architecture is identified, they can be effectively trained for the problem in hand.

\bibliographystyle{unsrtnat}
\bibliography{References}  

\begin{thebibliography}{43}
\providecommand{\natexlab}[1]{#1}
\providecommand{\url}[1]{\texttt{#1}}
\expandafter\ifx\csname urlstyle\endcsname\relax
  \providecommand{\doi}[1]{doi: #1}\else
  \providecommand{\doi}{doi: \begingroup \urlstyle{rm}\Url}\fi

\bibitem[HALL et~al.(1970)HALL, CALKIN, and SHOLAR]{hall1970}
BERT HALL, E~CALKIN, and M~SHOLAR.
\newblock Linear estimation of structural parameters from dynamic test data.
\newblock In \emph{11th Structures, Structural Dynamics, and Materials
  Conference}, page 1521, 1970.

\bibitem[Kabe(1985)]{kabe1985}
Alvar~M Kabe.
\newblock Stiffness matrix adjustment using mode data.
\newblock \emph{AIAA journal}, 23\penalty0 (9):\penalty0 1431--1436, 1985.

\bibitem[Ahmadian et~al.(1997)Ahmadian, Gladwell, and Ismail]{ahmadian1997}
H~Ahmadian, GML Gladwell, and F~Ismail.
\newblock Parameter selection strategies in finite element model updating.
\newblock 1997.

\bibitem[Friswell et~al.(1998)Friswell, Inman, and Pilkey]{friswell1998}
MI~Friswell, DJ~Inman, and Deborah~F Pilkey.
\newblock Direct updating of damping and stiffness matrices.
\newblock \emph{AIAA journal}, 36\penalty0 (3):\penalty0 491--493, 1998.

\bibitem[Yan et~al.(2017)Yan, Sun, and B{\"u}y{\"u}k{\"o}zt{\"u}rk]{yan2017}
Gang Yan, Hao Sun, and Oral B{\"u}y{\"u}k{\"o}zt{\"u}rk.
\newblock Impact load identification for composite structures using bayesian
  regularization and unscented kalman filter.
\newblock \emph{Structural Control and Health Monitoring}, 24\penalty0
  (5):\penalty0 e1910, 2017.

\bibitem[Zhang et~al.(2019)Zhang, Chen, Chen, Zheng,
  B{\"u}y{\"u}k{\"o}zt{\"u}rk, and Sun]{zhang2019}
Ruiyang Zhang, Zhao Chen, Su~Chen, Jingwei Zheng, Oral
  B{\"u}y{\"u}k{\"o}zt{\"u}rk, and Hao Sun.
\newblock Deep long short-term memory networks for nonlinear structural seismic
  response prediction.
\newblock \emph{Computers \& Structures}, 220:\penalty0 55--68, 2019.

\bibitem[Bishop(2006)]{Bishop2006}
Christopher~M. Bishop.
\newblock \emph{{Pattern Recognition and Machine Learning}}.
\newblock 2006.
\newblock ISBN 9780387310732.
\newblock URL \url{http://www.springer.com/us/book/9780387310732}.

\bibitem[LeCun et~al.(2015)LeCun, Bengio, and Hinton]{Lecun2015}
Yann LeCun, Yoshua Bengio, and Geoffrey Hinton.
\newblock {Deep learning}.
\newblock \emph{Nature}, 521\penalty0 (7553):\penalty0 436--444, 5 2015.
\newblock ISSN 0028-0836.
\newblock \doi{10.1038/nature14539}.
\newblock URL \url{http://www.nature.com/articles/nature14539}.

\bibitem[Goodfellow et~al.(2016)Goodfellow, Bengio, and
  Courville]{Goodfellow2016}
Ian Goodfellow, Yoshua Bengio, and Aaron Courville.
\newblock \emph{{Deep Learning}}.
\newblock 2016.
\newblock ISBN 9781405161251.
\newblock URL \url{https://www.deeplearningbook.org}.

\bibitem[Rafiei and Adeli(2017)]{Rafiei2017}
Mohammad~Hossein Rafiei and Hojjat Adeli.
\newblock {A novel machine learning-based algorithm to detect damage in
  high-rise building structures}.
\newblock \emph{Structural Design of Tall and Special Buildings}, 26\penalty0
  (18):\penalty0 1--11, 2017.
\newblock ISSN 15417808.
\newblock \doi{10.1002/tal.1400}.

\bibitem[Rouet-Leduc et~al.(2017)Rouet-Leduc, Hulbert, Lubbers, Barros,
  Humphreys, and Johnson]{Rouet-Leduc2017}
Bertrand Rouet-Leduc, Claudia Hulbert, Nicholas Lubbers, Kipton Barros,
  Colin~J. Humphreys, and Paul~A. Johnson.
\newblock {Machine Learning Predicts Laboratory Earthquakes}.
\newblock \emph{Geophysical Research Letters}, 44\penalty0 (18):\penalty0
  9276--9282, 2017.
\newblock ISSN 19448007.
\newblock \doi{10.1002/2017GL074677}.

\bibitem[Perol et~al.(2018)Perol, Gharbi, and
  Denolle]{Perol2018ConvolutionalLocation}
Thibaut Perol, Michaël Gharbi, and Marine Denolle.
\newblock {Convolutional neural network for earthquake detection and location}.
\newblock \emph{Science Advances}, 4\penalty0 (2):\penalty0 e1700578, 2 2018.
\newblock ISSN 2375-2548.
\newblock \doi{10.1126/sciadv.1700578}.
\newblock URL
  \url{https://advances.sciencemag.org/lookup/doi/10.1126/sciadv.1700578}.

\bibitem[Ross et~al.(2019)Ross, Trugman, Hauksson, and Shearer]{Ross2019}
Zachary~E. Ross, Daniel~T. Trugman, Egill Hauksson, and Peter~M. Shearer.
\newblock {Searching for hidden earthquakes in Southern California}.
\newblock \emph{Science}, 771\penalty0 (May):\penalty0 767--771, 2019.
\newblock ISSN 10959203.
\newblock \doi{10.1126/science.aaw6888}.

\bibitem[Pilania et~al.(2013)Pilania, Wang, Jiang, Rajasekaran, and
  Ramprasad]{Pilania2013}
Ghanshyam Pilania, Chenchen Wang, Xun Jiang, Sanguthevar Rajasekaran, and
  Ramamurthy Ramprasad.
\newblock {Accelerating materials property predictions using machine learning}.
\newblock \emph{Scientific Reports}, 3:\penalty0 1--6, 2013.
\newblock ISSN 20452322.
\newblock \doi{10.1038/srep02810}.

\bibitem[Liang et~al.(2018)Liang, Liu, Martin, and Sun]{Liang2018}
Liang Liang, Minliang Liu, Caitlin Martin, and Wei Sun.
\newblock {A deep learning approach to estimate stress distribution: a fast and
  accurate surrogate of finite-element analysis}.
\newblock \emph{Journal of the Royal Society Interface}, 15\penalty0 (138),
  2018.
\newblock ISSN 17425662.
\newblock \doi{10.1098/rsif.2017.0844}.

\bibitem[DeVries et~al.(2018)DeVries, Vi{\'{e}}gas, Wattenberg, and
  Meade]{DeVries2018}
Phoebe~M.R. DeVries, Fernanda Vi{\'{e}}gas, Martin Wattenberg, and Brendan~J.
  Meade.
\newblock {Deep learning of aftershock patterns following large earthquakes}.
\newblock \emph{Nature}, 560\penalty0 (7720):\penalty0 632--634, 2018.
\newblock ISSN 14764687.
\newblock \doi{10.1038/s41586-018-0438-y}.
\newblock URL \url{http://dx.doi.org/10.1038/s41586-018-0438-y}.

\bibitem[Wang and Sun(2018)]{Wang2018ALearning}
Kun Wang and Wai~Ching Sun.
\newblock {A multiscale multi-permeability poroplasticity model linked by
  recursive homogenizations and deep learning}.
\newblock \emph{Computer Methods in Applied Mechanics and Engineering},
  334:\penalty0 337--380, 2018.
\newblock ISSN 00457825.
\newblock \doi{10.1016/j.cma.2018.01.036}.

\bibitem[Kabacao{\u{g}}lu and Biros(2019)]{Kabacaoglu2019}
Gökberk Kabacao{\u{g}}lu and George Biros.
\newblock {Machine learning acceleration of simulations of Stokesian
  suspensions}.
\newblock \emph{Physical Review E}, 99\penalty0 (6):\penalty0 063313, 6 2019.
\newblock ISSN 2470-0045.
\newblock \doi{10.1103/PhysRevE.99.063313}.
\newblock URL \url{https://link.aps.org/doi/10.1103/PhysRevE.99.063313}.

\bibitem[Bergen et~al.(2019)Bergen, Johnson, de~Hoop, and Beroza]{Bergen2019}
Karianne~J.; Bergen, Paul~A.; Johnson, Maarten~V.; de~Hoop, and Gregory~C.
  Beroza.
\newblock {Machine learning for data-driven discovery in solid Earth
  geoscience}.
\newblock \emph{Science}, 363\penalty0 (6433):\penalty0 eaau0323, 3 2019.
\newblock ISSN 0036-8075.
\newblock \doi{10.1126/science.aau0323}.
\newblock URL
  \url{http://www.sciencemag.org/lookup/doi/10.1126/science.aau0323}.

\bibitem[Dana and Wheeler(2020)]{Dana2020}
Saumik Dana and Mary~F Wheeler.
\newblock {A machine learning accelerated FE{\$}{\^{}}2{\$} homogenization
  algorithm for elastic solids}.
\newblock 2020.
\newblock URL \url{http://arxiv.org/abs/2003.11372}.

\bibitem[Wagner and Rondinelli(2016)]{Wagner2016Theory-guidedScience}
Nicholas Wagner and James~M. Rondinelli.
\newblock {Theory-guided machine learning in materials science}.
\newblock \emph{Frontiers in Materials}, 3\penalty0 (June):\penalty0 1--9,
  2016.
\newblock ISSN 22968016.
\newblock \doi{10.3389/fmats.2016.00028}.

\bibitem[Rudy et~al.(2018)Rudy, Alla, Brunton, and Kutz]{Rudy2018}
Samuel Rudy, Alessandro Alla, Steven~L. Brunton, and J.~Nathan Kutz.
\newblock {Data-driven identification of parametric partial differential
  equations}.
\newblock 18\penalty0 (2):\penalty0 643--660, 2018.
\newblock URL \url{http://arxiv.org/abs/1806.00732}.

\bibitem[Weinan and Yu(2018)]{Weinan2018}
E.~Weinan and Bing Yu.
\newblock {The Deep Ritz Method: A Deep Learning-Based Numerical Algorithm for
  Solving Variational Problems}.
\newblock \emph{Communications in Mathematics and Statistics}, 6\penalty0
  (1):\penalty0 1--14, 2018.
\newblock ISSN 2194671X.
\newblock \doi{10.1007/s40304-018-0127-z}.

\bibitem[Raissi et~al.(2019)Raissi, Perdikaris, and Karniadakis]{Raissi2019}
M.~Raissi, P.~Perdikaris, and G.~E. Karniadakis.
\newblock {Physics-informed neural networks: A deep learning framework for
  solving forward and inverse problems involving nonlinear partial differential
  equations}.
\newblock \emph{Journal of Computational Physics}, 378:\penalty0 686--707,
  2019.
\newblock ISSN 10902716.
\newblock \doi{10.1016/j.jcp.2018.10.045}.
\newblock URL \url{https://doi.org/10.1016/j.jcp.2018.10.045}.

\bibitem[Kharazmi et~al.(2019)Kharazmi, Zhang, and Karniadakis]{Kharazmi2019}
E.~Kharazmi, Z.~Zhang, and G.~E. Karniadakis.
\newblock {Variational Physics-Informed Neural Networks For Solving Partial
  Differential Equations}.
\newblock pages 1--24, 2019.
\newblock URL \url{http://arxiv.org/abs/1912.00873}.

\bibitem[Fang and Zhan(2019)]{Fang2019AProblems}
Zhiwei Fang and Justin Zhan.
\newblock {A Physics-Informed Neural Network Framework For Partial Differential
  Equations on 3D Surfaces : Time Independent Problems}.
\newblock \emph{IEEE Access}, PP:\penalty0 1, 2019.
\newblock \doi{10.1109/ACCESS.2019.2963390}.

\bibitem[Haghighat et~al.(2020{\natexlab{a}})Haghighat, Raissi, Moure, Gomez,
  and Juanes]{Haghighat2020AMechanics}
Ehsan Haghighat, Maziar Raissi, Adrian Moure, Hector Gomez, and Ruben Juanes.
\newblock {A deep learning framework for solution and discovery in solid
  mechanics}.
\newblock 2020{\natexlab{a}}.

\bibitem[Mao et~al.(2020)Mao, Jagtap, and
  Karniadakis]{Mao2020Physics-informedFlows}
Zhiping Mao, Ameya~D. Jagtap, and George~Em Karniadakis.
\newblock {Physics-informed neural networks for high-speed flows}.
\newblock \emph{Computer Methods in Applied Mechanics and Engineering},
  360:\penalty0 112789, 2020.
\newblock ISSN 00457825.
\newblock \doi{10.1016/j.cma.2019.112789}.
\newblock URL \url{https://doi.org/10.1016/j.cma.2019.112789}.

\bibitem[Fuks and Tchelepi(2020)]{Fuks2020PhysicsMedia}
O~Fuks and H.~A. Tchelepi.
\newblock {Physics Based Deep Learning for Nonlinear Two-Phase Flow in Porous
  Media}.
\newblock In \emph{ECMOR XVII}, number Sep, pages 1--10. European Association
  of Geoscientists {\&} Engineers, 2020.
\newblock \doi{10.3997/2214-4609}.
\newblock URL
  \url{https://www.earthdoc.org/content/papers/10.3997/2214-4609.202035147}.

\bibitem[Sun et~al.(2020)Sun, Gao, Pan, and Wang]{Sun2020SurrogateData}
Luning Sun, Han Gao, Shaowu Pan, and Jian~Xun Wang.
\newblock {Surrogate modeling for fluid flows based on physics-constrained deep
  learning without simulation data}.
\newblock \emph{Computer Methods in Applied Mechanics and Engineering},
  361:\penalty0 112732, 2020.
\newblock ISSN 00457825.
\newblock \doi{10.1016/j.cma.2019.112732}.
\newblock URL \url{https://doi.org/10.1016/j.cma.2019.112732}.

\bibitem[Tartakovsky et~al.(2020)Tartakovsky, Marrero, Perdikaris, Tartakovsky,
  and Barajas-Solano]{Tartakovsky2020Physics-InformedProblems}
A.~M. Tartakovsky, C.~Ortiz Marrero, Paris Perdikaris, G.~D. Tartakovsky, and
  D.~Barajas-Solano.
\newblock {Physics-Informed Deep Neural Networks for Learning Parameters and
  Constitutive Relationships in Subsurface Flow Problems}.
\newblock \emph{Water Resources Research}, 56\penalty0 (5):\penalty0 1--16,
  2020.
\newblock ISSN 19447973.
\newblock \doi{10.1029/2019WR026731}.

\bibitem[Meng et~al.(2019)Meng, Li, Zhang, and Karniadakis]{Meng2019}
Xuhui Meng, Zhen Li, Dongkun Zhang, and George~Em Karniadakis.
\newblock {PPINN: Parareal Physics-Informed Neural Network for time-dependent
  PDEs}.
\newblock pages 1--17, 2019.
\newblock URL \url{http://arxiv.org/abs/1909.10145}.

\bibitem[Haghighat et~al.(2020{\natexlab{b}})Haghighat, Can, Madenci, and
  Juanes]{Haghighat2020AOperatorb}
Ehsan Haghighat, Ali Can, Erdogan Madenci, and Ruben Juanes.
\newblock {A nonlocal physics-informed deep learning framework using the
  peridynamic differential operator}.
\newblock \penalty0 (May), 2020{\natexlab{b}}.

\bibitem[Kharazmi et~al.(2020)Kharazmi, Zhang, and Karniadakis]{Kharazmi2020}
Ehsan Kharazmi, Zhongqiang Zhang, and George~Em Karniadakis.
\newblock {hp-VPINNs: Variational Physics-Informed Neural Networks With Domain
  Decomposition}.
\newblock pages 1--21, 2020.
\newblock URL \url{http://arxiv.org/abs/2003.05385}.

\bibitem[Jagtap et~al.(2020)Jagtap, Kharazmi, and Karniadakis]{Jagtap2020}
Ameya~D. Jagtap, Ehsan Kharazmi, and George~Em Karniadakis.
\newblock {Conservative physics-informed neural networks on discrete domains
  for conservation laws: Applications to forward and inverse problems}.
\newblock \emph{Computer Methods in Applied Mechanics and Engineering},
  365:\penalty0 113028, 2020.
\newblock ISSN 00457825.
\newblock \doi{10.1016/j.cma.2020.113028}.
\newblock URL \url{https://doi.org/10.1016/j.cma.2020.113028}.

\bibitem[Hornik et~al.(1989)Hornik, Stinchcombe, and White]{Hornik1989}
Kurt Hornik, Maxwell Stinchcombe, and Halbert White.
\newblock {Multilayer feedforward networks are universal approximators}.
\newblock \emph{Neural Networks}, 2\penalty0 (5):\penalty0 359--366, 1989.
\newblock ISSN 08936080.
\newblock \doi{10.1016/0893-6080(89)90020-8}.

\bibitem[G{\"{u}}ne¸ et~al.(2018)G{\"{u}}ne¸, Baydin, Pearlmutter, and
  Siskind]{Gune2018}
Atılım G{\"{u}}ne¸, Güne¸s Baydin, Barak~A Pearlmutter, and Jeffrey~Mark
  Siskind.
\newblock {Automatic Differentiation in Machine Learning: a Survey}.
\newblock \emph{Journal of Machine Learning Research}, 18:\penalty0 1--43,
  2018.
\newblock URL \url{http://www.jmlr.org/papers/volume18/17-468/17-468.pdf}.

\bibitem[Nocedal and Wright(2006)]{Robinson2006}
Jorge Nocedal and Stephen~J. Wright.
\newblock \emph{{Numerical Optimization}}, volume~17 of \emph{Springer Series
  in Operations Research and Financial Engineering}.
\newblock Springer New York, 2006.
\newblock ISBN 978-0-387-30303-1.
\newblock \doi{10.1007/978-0-387-40065-5}.
\newblock URL \url{http://link.springer.com/10.1007/978-0-387-40065-5}.

\bibitem[Kingma and Ba(2015)]{Kingma2015Adam:Optimization}
Diederik~P. Kingma and Jimmy~Lei Ba.
\newblock {Adam: A method for stochastic optimization}.
\newblock \emph{3rd International Conference on Learning Representations, ICLR
  2015 - Conference Track Proceedings}, pages 1--15, 2015.

\bibitem[Sun(2019)]{Sun2019OptimizationAlgorithms}
Ruoyu Sun.
\newblock {Optimization for deep learning: theory and algorithms}.
\newblock \emph{arXiv}, pages 1--60, 2019.

\bibitem[Kim and De~Weck(2005)]{Kim2005AdaptiveGeneration}
I.~Y. Kim and O.~L. De~Weck.
\newblock {Adaptive weighted-sum method for bi-objective optimization: Pareto
  front generation}.
\newblock \emph{Structural and Multidisciplinary Optimization}, 29\penalty0
  (2):\penalty0 149--158, 2005.
\newblock ISSN 1615147X.
\newblock \doi{10.1007/s00158-004-0465-1}.

\bibitem[Wang et~al.(2020)Wang, Teng, and
  Perdikaris]{Wang2020UnderstandingNetworks}
Sifan Wang, Yujun Teng, and Paris Perdikaris.
\newblock {Understanding and mitigating gradient pathologies in
  physics-informed neural networks}.
\newblock pages 1--28, 2020.
\newblock URL \url{http://arxiv.org/abs/2001.04536}.

\bibitem[Haghighat and Juanes(2019)]{Haghighat2019SciANN:Networks}
Ehsan Haghighat and Ruben Juanes.
\newblock {SciANN: A Keras wrapper for scientific computations and
  physics-informed deep learning using artificial neural networks}.
\newblock 2019.

\end{thebibliography}






\end{document}